\documentclass[10pt,twocolumn,letterpaper]{article}

\usepackage[pagenumbers]{cvpr} 

\definecolor{cvprblue}{rgb}{0.21,0.49,0.74}
\usepackage[pagebackref,breaklinks,colorlinks,allcolors=cvprblue]{hyperref}

\usepackage[utf8]{inputenc} 
\usepackage[T1]{fontenc}    
\usepackage{hyperref}       
\usepackage{url}            
\usepackage{booktabs}       
\usepackage{amsfonts}       
\usepackage{nicefrac}       
\usepackage{microtype}      
\usepackage[table]{xcolor}

\usepackage{graphicx}
\usepackage{amsmath}
\usepackage{amssymb}

\usepackage{listings}
\usepackage{mwe} 
\usepackage{makecell}
\usepackage{color, colortbl}
\usepackage[normalem]{ulem} 
\usepackage{algorithm}
\usepackage{algorithmic}

\usepackage{wrapfig}
\usepackage{caption}
\usepackage{pifont} 
\usepackage{multirow}
\usepackage{booktabs}
\usepackage{float}

\usepackage{chngpage}
\usepackage{wrapfig}
\usepackage{adjustbox}
\usepackage{comment}
\usepackage{color}

\usepackage{listings}

\usepackage{bm}
\usepackage[capitalize]{cleveref}

\usepackage[toc]{appendix}
\usepackage{etoc}
\usepackage{minitoc}
\usepackage{mathtools}

\usepackage{amsthm}

\theoremstyle{definition}

\theoremstyle{remark}

\crefname{section}{Sec.}{Secs.}
\Crefname{section}{Section}{Sections}
\Crefname{table}{Table}{Tables}
\crefname{table}{Tab.}{Tabs.}
\etocsettocstyle{\section*{Appendix}}{}

\definecolor{Gray}{gray}{0.9}
\definecolor{ImportantColor}{rgb}{0.63, 0.79, 0.95}
\newcolumntype{g}{>{\columncolor{ImportantColor}}c}
\newcolumntype{?}{!{\vrule width 1pt}}
\newcommand{\tb}[3]{\setlength{\tabcolsep}{#2mm}\begin{tabular}{#1}#3\end{tabular}}

\graphicspath{
{figs/},
}

\newcommand*{\belowrulesepcolor}[1]{%
  \noalign{%
    \kern-\belowrulesep 
    \begingroup 
      \color{#1}%
      \hrule height\belowrulesep 
    \endgroup 
  }%
} 
\newcommand*{\aboverulesepcolor}[1]{%
  \noalign{%
    \begingroup 
      \color{#1}%
      \hrule height\aboverulesep 
    \endgroup 
    \kern-\aboverulesep 
  }%
}

\definecolor{codegreen}{rgb}{0,0.6,0}
\definecolor{codegray}{rgb}{0.5,0.5,0.5}
\definecolor{codepurple}{rgb}{0.58,0,0.82}
\definecolor{backcolour}{rgb}{0.95,0.95,0.92}

\lstdefinestyle{mystyle}{
    backgroundcolor=\color{backcolour},   
    commentstyle=\color{codegreen},
    keywordstyle=\color{magenta},
    numberstyle=\tiny\color{codegray},
    stringstyle=\color{codepurple},
    basicstyle=\ttfamily\footnotesize,
    breakatwhitespace=false,         
    breaklines=true,                 
    captionpos=b,                    
    keepspaces=true,                 
    numbers=left,                    
    numbersep=5pt,                  
    showspaces=false,                
    showstringspaces=false,
    showtabs=false,                  
    tabsize=2
}

\def\paperID{45643} 
\def\confName{CVPR}
\def\confYear{2026}

\usepackage{amsmath,amsfonts,bm}

\def\eqref#1{equation~\ref{#1}}

\def\1{\bm{1}}

\DeclareMathAlphabet{\mathsfit}{\encodingdefault}{\sfdefault}{m}{sl}
\SetMathAlphabet{\mathsfit}{bold}{\encodingdefault}{\sfdefault}{bx}{n}

\newcommand{\arginf}{\mathop{\mathrm{arginf}}}

\newcommand{\method}{COT-FM}
\newcommand{\methodfull}{Cluster-wise Optimal Transport Flow Matching}

\title{\method{}: \methodfull{}}

\author{
Chiensheng Chiang$^{*}$\qquad
Kuan-Hsun Tu$^{*\dagger}$\qquad
Jia-Wei Liao$^{*}$\\[6pt]
Cheng-Fu Chou\qquad
Tsung-Wei Ke\\[6pt]
National Taiwan University\\[4pt]
}

\begin{document}
\twocolumn[{
    \renewcommand\twocolumn[1][]{#1}%
    \maketitle   
    \begin{center}
        \newcommand{\teaserwidth}{\textwidth}
        \vspace{-0.15in}
        \centerline{
        \includegraphics[width=\teaserwidth, clip]{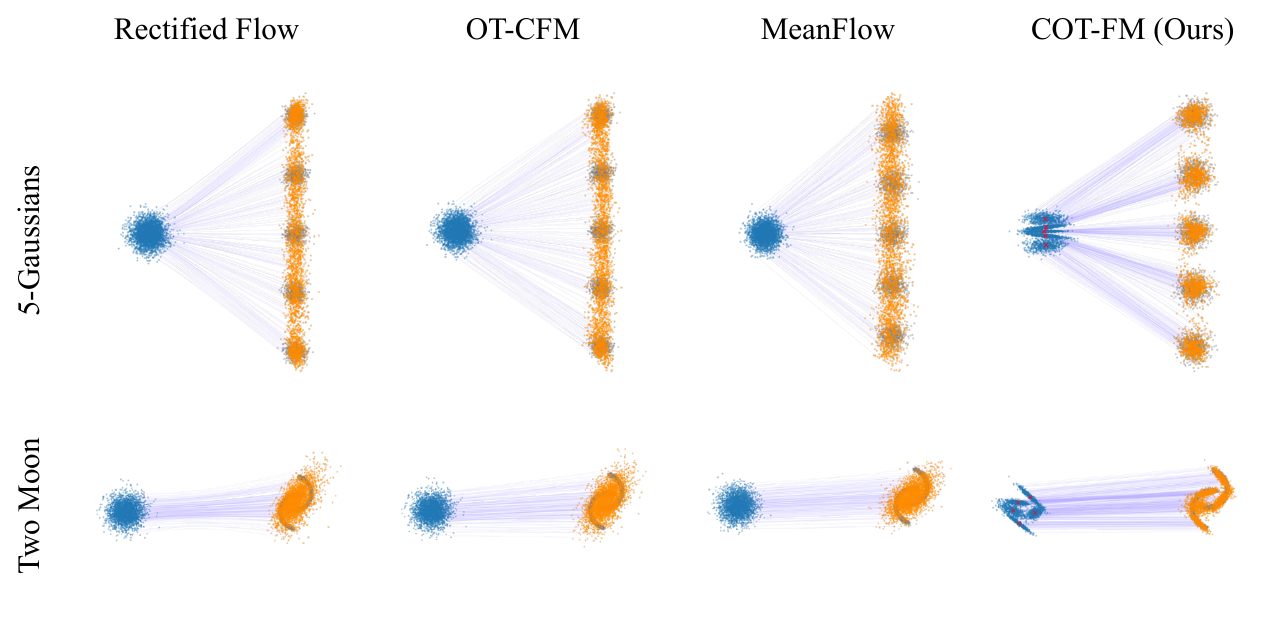}
        }
        \vspace{-5ex}
        \captionof{figure}{\textbf{\method{} Yields Straight, Structure-Preserving Transport Flows.} Blue points show the source distribution (Gaussian), and gray points show the target distributions (a 5-component Gaussian mixture and Two Moons). Orange points are generated samples, and purple lines denote the learned transport trajectories. Red crosses mark cluster means under our Cluster-wise Optimal Transport Flow Matching (COT-FM). The method yields straight trajectories while still capturing the structure of each target distribution.}
        
    \label{fig:teaser}
    \vspace{-8pt}
    \end{center}
}]

\begin{abstract}
\vspace{-50pt}

We introduce \method{}, a general framework that reshapes the probability path in Flow Matching (FM) to achieve faster and more reliable generation. FM models often produce curved trajectories due to random or batch-wise couplings, which increase discretization error and reduce sample quality. \method{} fixes this by clustering target samples and assigning each cluster a dedicated source distribution obtained by reversing pretrained FM models. This divide-and-conquer strategy yields more accurate local transport and significantly straighter vector fields, all without changing the model architecture. As a plug-and-play approach, \method{} consistently accelerates sampling and improves generation quality across 2D datasets, image generation benchmarks, and robotic manipulation tasks.

\end{abstract}

\section{Introduction}
\renewcommand{\thefootnote}{}
\footnotetext{* Equal contribution. $\dagger$ Corresponding author. \\
\phantom{*} Project page: \url{https://embodiedai-ntu.github.io/cotfm}}
\renewcommand{\thefootnote}{\arabic{footnote}}
\setcounter{footnote}{0}

Generative modeling seeks to learn a transformation that maps a simple, known source distribution to the complex, partially known data distribution~\cite{kingma2013auto, rezende2015variational, sohl2015deep, song2019generative, ho2020denoising, song2020score}, for generating new data samples.  Flow Matching (FM)~\cite{lipman2022flow,albergo2022building} is a specific framework: it regresses a deterministic vector field inducing the desired probability path between two distributions.  During inference, source samples are transformed into data samples through integration along the learned vector field.  Due to its flexibility and scalability, FM recently emerges as an effective alternative to generative models, demonstrating promising results in a wide range of tasks~\cite{tong2020trajectorynet,liu2023instaflow,black2024pi_0}.

The formulation of FM is general.  It encompasses different types of generative models, such as diffusion models~\cite{song2020score,ho2020denoising}, if the target vector field is constructed appropriately~\cite{lipman2022flow}.  In practice, straight vector fields are preferred over curved ones, as straighter paths incur lower time-discretization error and therefore reduce sampling steps for generation.  Meanwhile, since FM models often do not learn the entire vector field during training due to computational limitations\footnote{For D-dimensional Gaussian distributions, $e^{\alpha D}$ samples are required to cover the entire vector field~\cite{vershynin2018high}.}, time-discretization error may move samples to unseen locations, leading to distorted transport and low-quality generation.  To enforce straightness, one can construct the vector field based on the optimal transport (OT) map~\cite{benamou2000computational}--the optimal couplings of samples between two distributions with minimal transport cost.  However, the exact solution of OT map is computationally inefficient\footnote{The OT map requires cubic time and quadratic memory complexity in the number of samples.} to obtain for large datasets~\cite{cuturi2013sinkhorn,tong2020trajectorynet}.

To address this limitation, most FM models either adopt random, independent couplings~\cite{lipman2022flow} or approximate the global optimal couplings with batch optimal couplings~\cite{fatras2021minibatchoptimaltransportdistances, nguyen2022improvingminibatchoptimaltransport, pooladian2023multisample,kornilov2024optimal, haxholli2024minibatchoptimaltransportperplexity,davtyan2025faster,lin2025beyond, cheng2025curse}, to construct the vector field during training.  While conceptually simple, the former yield frequent path crossings and the latter struggle with locality of batchwise approximations~\cite{fatras2021unbalanced}, both resulting in curved paths.  To straighten the learned paths, $k$-Rectified Flow~\cite{liu2022flow} proposes to iteratively optimize FM models with couplings between source and generated samples.  Although this approach provably enhances straightness, it repeatedly trains FM models on self-generated samples, leading to model collapse and therefore degrading generation quality over time~\cite{zhu2025analyzingmitigatingmodelcollapse}.

Another line of work bypasses the challenge of learning straight paths, attempting to distill from multi-step model ~\cite{salimans2022progressive, geng2023one, sauer2024adversarial, song2023consistency, yin2024one, zhou2024score} or to learn the average vector field ~\cite{kimconsistency, frans2024one,boffi2024flow, geng2025mean} for accelerating generation.  The latter idea is to skip sampling steps during inference with the average vector field.  Despite their efficiency, these approaches do not modulate the underlying instant vector field of FM models, which remains curved due to random coupling strategy.  As a result, such shortcut methods only reduce step count, but do not enhance generation quality. To illustrate these issues, we present a toy example in Fig.~\ref{fig:teaser}, where the data distribution comprises five modes and the source distribution is Gaussian.  We have three observations: (1) training FM models with either random or  batch optimal couplings like OT-CFM~\cite{tong2024improvinggeneralizingflowbasedgenerative} results in curved flows, (2) shortcut approaches like MeanFlow~\cite{geng2025mean} do not straighten the learned flows, and (3) curved paths lead to distorted data generation. 

We introduce \methodfull{} (\method{}), a general and effective framework that accelerates and enhances a wide range of FM models. We observe that training data of existing generative modeling tasks naturally comprise multiple modes, while data of the same mode can be generated from similar source samples, as shown in Fig.~\ref{fig:teaser}. Based on these observations, our key insight is to partition data samples into clusters, each assigned its own source distribution, rather than mapping the entire data distribution from a single source. This idea brings two advantages: (1) it reduces each optimal coupling problem to a smaller cluster-level match, which reduces the number of data–source samples and makes batch optimal couplings more effective; and (2) by restricting the space of source distributions, learning the overall vector field becomes more efficient. In order to find optimal source distributions for each data cluster, our second insight is to bootstrap from pre-trained FM models. Since paths learned by these models are naturally reversible and non-intersecting, we can easily obtain source distributions for each cluster, incurring less frequent crossings, by reversing the sampling process of these trained models.

Specifically, \method{} alternates optimization of the target vector field and the FM model. In the first stage, the FM model regresses the target vector field derived from a prior or previously estimated cluster-wise source distributions. In the second stage, we update these source distributions by reversing the learned paths for each data sample and computing their Gaussian statistics, followed by approximating OT within each cluster using batch-wise couplings. Notably, our formulation accommodates different types of clustering, including class labels or textual descriptions for conditional generation and unsupervised clustering for unconditional generation. Moreover, \method{} only modulates the target probability path, without altering the FM architecture or input–output mechanisms, making it compatible with most existing FM models and able to improve both generation speed and quality.

\method{} shows strong empirical gains in both one-step and few-step generation. On 2D transport benchmarks, it achieves the best results across all methods, reducing Wasserstein distance from 0.5421 to 0.1995 on Mixture of 5-Gaussians, from 0.1006 to 0.0266 on Two Moons, and from 0.3900 to 0.2550 on Checkerboard, while also attaining the lowest curvature. On CIFAR-10~\cite{krizhevsky2009learning}, \method{} improves Rectified Flow~\cite{liu2022flow} from 12.6 to 8.23 FID at 10 steps and from 4.45 to 3.97 at 50 steps. At very low sampling budgets, it also reduces 1-step FID from 378.0 to 205.0 and 2-step FID from 173 to 59.1. It also enhances MeanFlow~\cite{geng2025mean}, lowering FID from 2.92 to 2.60 (1-step) and 2.88 to 2.53 (2-step). On LIBERO robotic manipulation~\cite{liu2023libero}, \method{} reaches 96.1\% (Spatial) and 94.5\% (Long) with just 1 NFE, while the FLOWER policy~\cite{reuss2025flower} requires 4 NFEs to reach 97.1\% and 93.5\%. Together, these results show that clustering targets and learning cluster-wise source distributions lead to straighter transport paths and more reliable low-step generation. These consistent improvements across domains confirm that reducing flow curvature is key to enabling high-quality generation under extremely low sampling budgets.

\begin{figure}[t!]
    \centering
    \includegraphics[width=\linewidth]{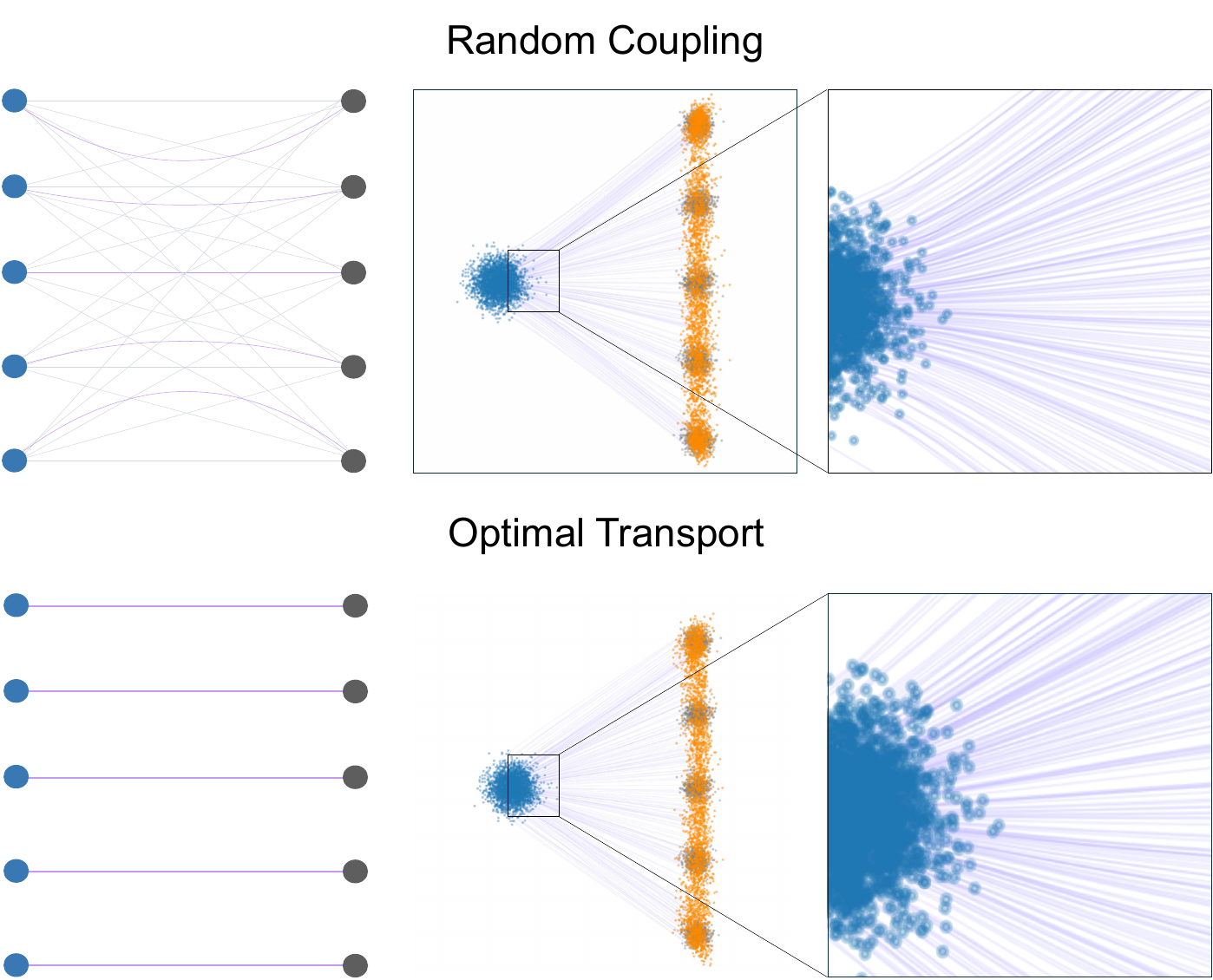}
    \caption{\textbf{Vector field results from different coupling strategies.} Random coupling forces the model to regress inconsistent (gray) velocity targets, creating ambiguous intersections and pushing the model toward an averaged (purple) direction, which produces curved velocity fields. In contrast, optimal transport provides consistent couplings with fewer intersections, enabling the model to learn much straighter velocity fields.}
    \label{fig:curvature}
\end{figure}

\begin{figure*}[t!]
    \centering
    \includegraphics[width=\textwidth]{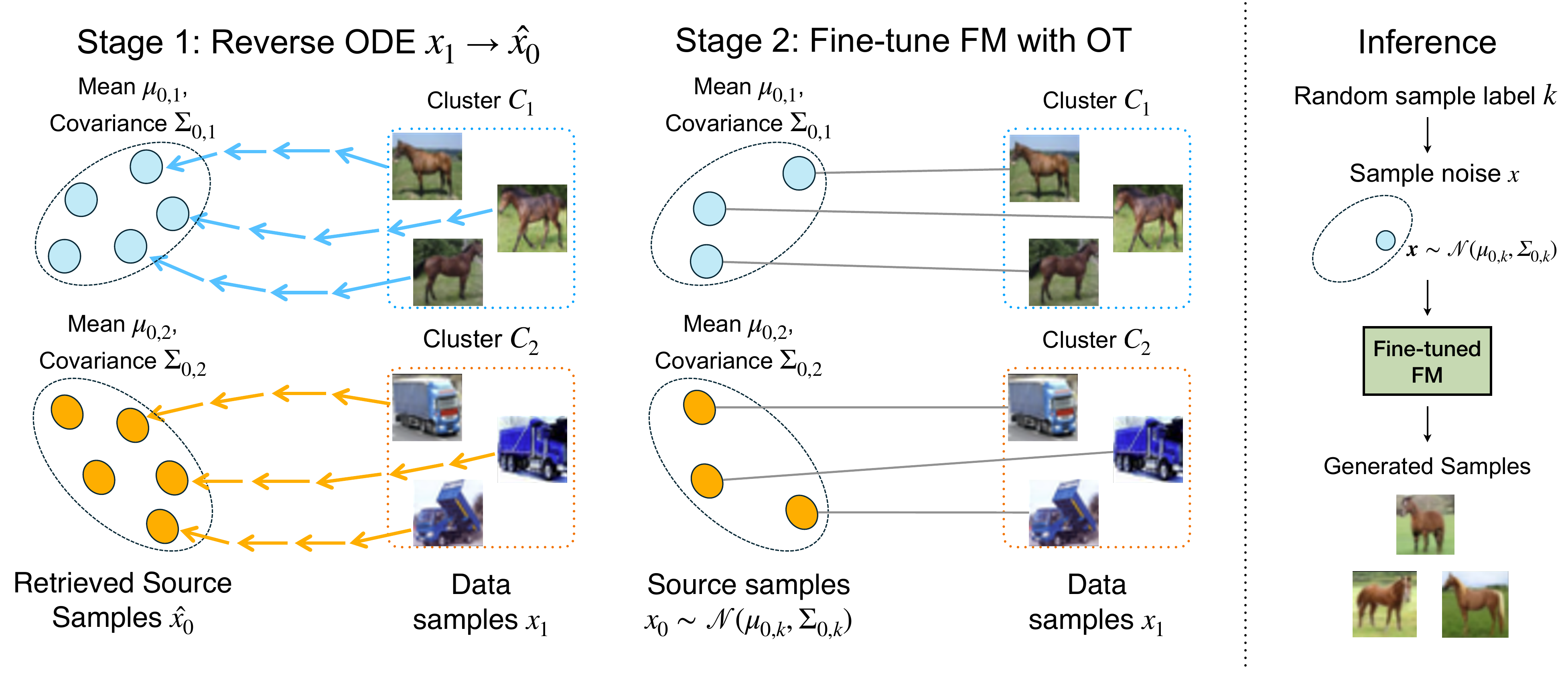}
    \caption{\textbf{Overview of our proposed method.} In Stage 1, we cluster the dataset, and for each image within a cluster $\mathcal{C}_k$, we reverse the ODE back to the noise space using a pretrained flow model to compute the new mean $\bm{\mu}_{0,k}$ and covariance $\bm{\Sigma}_{0,k}$ for that cluster. In Stage 2, we apply optimal transport to finetune the pretrained flow model across all clusters, encouraging it to align with the distributional structure captured in Stage 1. During inference, we first sample a cluster index $k$, then draw a noise $\bm{x}$ from $\mathcal{N}(\bm{\mu}_{0,k}, \bm{\Sigma}_{0,k})$, and use this noise to generate an image through the finetuned flow model.}
    \label{fig:method}
\end{figure*}

\section{Preliminary of Flow Matching}
\label{sec:preliminary}

The objective of flow matching (FM) models is to match a time-dependent vector field transporting samples from a source distribution $p_0$ to those of a target distribution $p_1$.  This formulation is conceptually simple and computationally efficient: it avoids expensive simulation during training while enabling generation through ordinary differential equation (ODE) integration during testing.

Formally, let $\bm{v}_t: [0, 1] \times \mathbb{R}^d    \rightarrow \mathbb{R}^d$ denote a time-varying \textit{vector field} that defines an ODE: $\mathrm{d}\bm{x}_t = \bm{v}_t(\bm{x}_t) \, \mathrm{d}t$.  The integration map (or \textit{flow}) of the ODE describes the transformation of sample $\bm{x}_0$ along the vector field from time $0$ to $t$.  In other words, the flow reshapes a simple source distribution $p_0$ to a complex target distribution $p_1$, where $p_t$ denote the \textit{probability path}--intermediate density functions transported along the vector field from time $0$ to $t$.  The FM objective regresses the vector field $\bm{v}_t$ with a neural network $\bm{v}_\theta$ parameterized by weights $\theta$, which is formulated as:
\begin{equation}
    \mathcal{L}_\textsf{FM}(\theta) = \mathbb{E}_{t, \bm{x}_t \sim p_t} \|\bm{v}_\theta(\bm{x}_t, t) - \bm{v}_t(\bm{x}_t)\|_2^2.
\end{equation}

However, the derivation of vector field $\bm{v}_t$ and probability path $p_t$ is often intractable for general source and target distributions~\cite{tong2024improvinggeneralizingflowbasedgenerative}.  To address this limitation, \cite{tong2024improvinggeneralizingflowbasedgenerative,lipman2022flow} suggest to construct the target probability path via a mixture of simpler conditional probability paths $p_t(\bm{x}_t | \bm{z})$ with variable $\bm{z}$, which may flexibly correspond to either data sample $\bm{x}_1$ or a pair of source-data sample $(\bm{x}_0, \bm{x}_1)$.  Marginalizing $p_t(\bm{x}_t | \bm{z})$ over distribution $q(\bm{z})$ produces the marginal probability path:
\begin{align}
    p_t(\bm{x}_t) = \int p_t(\bm{x}_t | \bm{z})q(\bm{z}) \, \mathrm{d} \bm{z}.
\label{eqn:marginal_probpath}
\end{align}
Meanwhile, we define the marginal vector field: 
\begin{align}
    \bm{v}_t(\bm{x}_t) = \mathbb{E}_{q(\bm{z})} \left[ \frac{\bm{v}_t(\bm{x}_t|\bm{z})p_t(\bm{x}_t|\bm{z})}{p_t(\bm{x}_t)} \right].
\label{eqn:marginal_vecfield}
\end{align}
If conditional vector field $\bm{v}_t(\bm{x}_t|\bm{z})$ generates conditional probability path $p_t(\bm{x}_t|\bm{z})$, the marginal vector field $\bm{v}_t(\bm{x}_t)$ is shown to generate the marginal probability path $p_t(\bm{x}_t)$~\cite{tong2024improvinggeneralizingflowbasedgenerative}.  Lastly, we define the conditional flow matching (CFM) objective as:
\begin{align}
    \mathcal{L}_\textsf{CFM}(\theta) = \mathbb{E}_{t, q(\bm{z}), p_t(\bm{x}_t|\bm{z})} \|\bm{v}_\theta(\bm{x}_t, t) - \bm{v}_t(\bm{x}_t | \bm{z})\|_2^2
    \label{eqn:cfm_loss}
\end{align}
leading to identical gradients as FM loss:
\begin{align}
    \nabla_\theta \mathcal{L}_\textsf{FM}(\theta)=\nabla_\theta \mathcal{L}_\textsf{CFM}(\theta).
\end{align}

\paragraph{Random Coupling.}
Randomly pairing a source sample $\bm{x}_0 \sim p_0$ with a target sample $\bm{x}_1 \sim p_1$ is a common way for training FM models.  Specifically, random coupling is formulated by setting variable $\bm{z}$ to a pair of source-target sample $(\bm{x}_0, \bm{x}_1)$, the distribution $q(\bm{z}) = p_0(\bm{x}_0)p_1(\bm{x}_1)$, and the conditional vector field $\bm{v}_t(\bm{x}_t|\bm{z}) = \bm{x}_1 - \bm{x}_0$. While each sample pair induces a straight path, the marginal vector field is curved as it aggregates paths of different sample pairs at each point $\bm{x}$ and time $t$. As illustrated in Fig.~\ref{fig:curvature}, this averaging of conflicting directions produces curved trajectories that lead to distribution misalignment.

\paragraph{Optimal Coupling.}
To enforce straightness, one can set distribution $q(z)$ to be the 2-Wasserstein OT map:
\begin{align}
    \pi(\bm{x}_0, \bm{x}_1) = \arginf_{\pi \in \Pi} \int \|\bm{x}_0 - \bm{x}_1\|_2^2 \, \mathrm{d} \pi(\bm{x}_0, \bm{x}_1)
    \label{eqn:ot}
\end{align}

where $\Pi$ denotes the set of all possible transport plans.  However, calculation of OT map $\pi(\bm{x}_0, \bm{x}_1)$ requires cubic time complexity in the number of samples, while continuous source distributions like Gaussian have infinite number of samples, therefore solving global OT map for large datasets is computationally infeasible.  To address this limitation, existing approaches often approximate the OT map with batch-wise OT map~\cite{fatras2021minibatchoptimaltransportdistances, nguyen2022improvingminibatchoptimaltransport, pooladian2023multisample,kornilov2024optimal, haxholli2024minibatchoptimaltransportperplexity,davtyan2025faster,lin2025beyond, cheng2025curse}.  However, with limited size of mini-batches, these methods struggle with locality of batchwise approximations~\cite{fatras2021unbalanced}, resulting in curved paths. For a more comprehensive discussion of related work, please refer to the Supplementary Material.
\section{Method}
\label{sec:method}

We propose \method{}, a general, plug-and-play framework that seeks an optimal vector field to accelerate and enhance generation quality of FM models.  The core idea is to divide-and-conquer the calculation of optimal couplings by clustering target samples, identifying corresponding source distributions for each cluster, and approximately solving optimal couplings within each cluster.  The FM model is then updated by regressing this vector field with straighter flows.  \method{} alternates between optimizing the target vector field and the FM model.  Notably, \method{} only modulates the underlying target probability path of the FM model, without modifying its architecture or input-output mechanisms.  This design makes \method{} broadly applicable across diverse FM models.  An overview of the proposed framework is illustrated in Fig.~\ref{fig:method}.

\subsection{Constructing Cluster-wise Target Vector Field}
\label{sec:clustering}

Our \method{} begins by partitioning target samples into clusters, then identifies source distributions and constructs a local target vector field for each cluster.  This formulation is efficient and general: it reduces the number of source-target samples, making approximation of OT more effective, and it generalizes to different types of clustering, such as class labels or textual descriptions for text-conditional generation and unsupervised clustering for unconditional generation.

\paragraph{Identifying Cluster-wise Source Distributions.}
To search individual source distributions, we propose to bootstrap from pre-trained FM models. Although originally trained with random coupling, their learned flows are naturally reversible while non-intersecting.  The former property allows us to estimate cluster-wise source distributions by integrating the flow backward to trace source samples that generate every data sample of a cluster; the latter property ensures paths between different clusters have few crossings.
Formally, we reverse ODE integration to retrieve the source sample of data sample $\bm{x}_{1}$:
\begin{align}
    \hat{\bm{x}}_{0} := \bm{x}_{1} - \int_0^1 \bm{v}_\theta(\hat{\bm{x}}_t, t) \, \mathrm{d}t,
    \label{eq:reverse_ode}
\end{align}
where $\bm{v}_\theta$ is the velocity field from a trained FM model and $\hat{\bm{x}}_t$ denotes reversely transported sample $\bm{x}_1$ from time $1$ to $t$.  We denote cluster $\mathcal{C}_k$ as a set of data samples whose clustering index is $k$.  Given $K$ clusters $\{\mathcal{C}_k\}_{k=1}^K$, we obtain source samples $\hat{X}_{0, k} = \{\hat{\bm{x}}_0 | \bm{x}_1 \in \mathcal{C}_k\}$ of every data sample within cluster $\mathcal{C}_k$.  We approximate their source distribution $p_{0, k}$ as Gaussian based on estimated mean and covariance:
\begin{align}
&\bm{\mu}_{0, k} = \frac{1}{|\hat{X}_{0, k}|}\sum_{\hat{\bm{x}}_0 \in \hat{X}_{0, k}} \hat{\bm{x}}_0, \label{eq:cluster_mean}\\
&\bm{\Sigma}_{0, k} = \frac{1}{|\hat{X}_{0, k}|} \sum_{\hat{\bm{x}}_0 \in \hat{X}_{0, k}} (\hat{\bm{x}}_0 - \bm{\mu}_{0, k})(\hat{\bm{x}}_0 - \bm{\mu}_{0, k})^\top, \\
&p_{0, k}(\bm{x}) = \mathcal{N}(\bm{x}; \bm{\mu}_{0, k}, \bm{\Sigma}_{0, k}).
\label{eq:cluster_cov}
\end{align}

\paragraph{Extension to Non-fixed Clustering.}
So far, identifying cluster-wise source distributions assumes a fixed set of clusters, which may not hold in some applications. For example, in vision-conditioned robot policies, treating each observation as a cluster causes the number of clusters to grow during rollout. To address this, we introduce a learning-based module that predicts the source distribution for each cluster. See the Supplementary Material for additional details.

\begin{algorithm}[t]
\caption{Constructing the cluster-wise vector field}
\label{algo:init-sources}
\begin{algorithmic}[1]
\STATE \textbf{Input:} Trained FM model $\bm{v}_\theta$ and $K$ clusters $\{\mathcal{C}_k\}_{k=1}^K$.
\FOR{$k = 1$ to $K$}
    \STATE Retrieve source samples with reverse ODE:
    \STATE $\quad \hat{X}_{0, k} \leftarrow \{\bm{x}_{1} - \int_0^1 \bm{v}_\theta(\hat{\bm{x}}_t, t) \, \mathrm{d}t | \bm{x}_{1} \in \mathcal{C}_k\}$
    \STATE Estimate source distributions:
    \STATE $\quad n_k \leftarrow |\hat{X}_{0, k}|$
    \STATE $\quad \bm{\mu}_{0, k} \leftarrow \frac{1}{n_k}\sum_{\hat{\bm{x}}_0 \in \hat{X}_{0, k}} \hat{\bm{x}}_0$
    \STATE $\quad \bm{\Sigma}_{0, k} \leftarrow \frac{1}{n_k} \sum_{\hat{\bm{x}}_0 \in \hat{X}_{0, k}} (\hat{\bm{x}}_0 - \bm{\mu}_{0, k})(\hat{\bm{x}}_0 - \bm{\mu}_{0, k})^\top$
    \STATE $\quad p_{0, k}(\bm{x}) \leftarrow \mathcal{N}(\bm{x}; \bm{\mu}_{0, k}, \bm{\Sigma}_{0, k})$
    \STATE Draw source samples $X_{0,k} \leftarrow \{\bm{x}_{0} \mid \bm{x}_{0} \sim p_{0,k}\}$
    \STATE Cluster-wise OT map: $\pi_k \leftarrow \text{OT}(X_{0,k}, \mathcal{C}_k)$
\ENDFOR
\STATE \textbf{Output:} Cluster-wise source distributions $\{p_{0,k}\}_{k=1}^K$ and OT maps $\{\pi_{k}\}_{k=1}^K$.
\end{algorithmic}
\end{algorithm}

\paragraph{Approximating OT within Each Cluster.}
To enforce straight paths, \method{} next updates the target vector field by calculating separate OT maps between each cluster of data samples and their assigned source distributions.  Although exact solution of cluster-wise OT maps remains computationally intractable, our method reduces the number of source-target samples, thereby making batch-wise approximation more feasible.  Specifically, for cluster $\mathcal{C}_k$, we draw a set of source samples $X_{0,k} = \{\bm{x}_{0} \mid \bm{x}_{0} \sim p_{0,k}\}$ from estimated source distributions $p_{0,k}$ and calculate the OT map $\pi_k$ between source samples $X_{0,k}$ and data samples $\mathcal{C}_k$ based on Eqn.~\ref{eqn:ot}.  These cluster-wise OT maps $\{\pi_k\}_{k=1}^K$ jointly, inherently composes a vector field (Eqn.~\ref{eqn:marginal_vecfield}), which the FM model is later optimized to regress.  The detailed procedure of constructing cluster-wise vector field is presented in Algorithm~\ref{algo:init-sources}.

\begin{algorithm}[t]
\caption{Optimizing the FM model}
\label{algo:train-fm}
\begin{algorithmic}[1]
\STATE \textbf{Input:} FM model $\bm{v}_\theta$, batch size $M$, training steps $L$, $K$ clusters $\{\mathcal{C}_k\}_{k=1}^K$, and cluster-wise OT maps $\{\pi_{k}\}_{k=1}^K$.
\FOR{$l = 1$ to $L$}
    \STATE Initialize minibatch: $B \leftarrow \{\}$
    \FOR{$m = 1$ to $M$}
        \STATE Sample cluster $k$ with probability $\frac{|\mathcal{C}_k|}{\sum_{j=1}^K |\mathcal{C}_j|}$
        \STATE Sample a source-target pair: $(\bm{x}_{0}, \bm{x}_{1}) \sim \pi_k$
        \STATE Append to minibatch: $B \leftarrow B \cup \{(\bm{x}_{0}, \bm{x}_{1})\}$
    \ENDFOR
    \STATE $\mathcal{L}_{\textsf{CFM}}(\theta) = \mathbb{E}_{t, (\bm{x}_0, \bm{x}_1) \sim B} \| \bm{v}_\theta(\bm{x}_{t}, t) - (\bm{x}_{1} - \bm{x}_{0}) \|_2^2$
    \STATE $\theta \leftarrow \theta - \gamma \nabla_\theta \mathcal{L}_{\textsf{CFM}}(\theta)$
\ENDFOR
\STATE \textbf{Output:} Updated FM model $\bm{v}_\theta$.
\end{algorithmic}
\end{algorithm}

\begin{algorithm}[t]
\caption{Sampling}
\label{algo:sample}
\begin{algorithmic}[1] 
\STATE \textbf{Input:} Cluster-wise source distributions $\{p_{0, k}\}_{k=1}^K$, FM model $\bm{v}_\theta$, number of sampling steps $T$.
\STATE Sample a cluster index $k$ (as in Algorithm~\ref{algo:train-fm}, line 5)
\STATE Draw source sample: $\bm{x} \sim p_{0, k}$
\FOR{$t=0$ to $T-1$}
    \STATE $\bm{x} \leftarrow \bm{x} + \bm{v}_\theta(\bm{x}, \frac{t}{T}) \cdot \frac{1}{T}$
\ENDFOR
\STATE \textbf{Output:} Generated target sample $\bm{x}$.
\end{algorithmic}
\end{algorithm}

\begin{table*}[t]
    \centering
    \resizebox{\linewidth}{!}{
    \begin{tabular}{l|c|cc|cc|cc}
        \toprule
        \multirow{2.5}{*}{Dataset: 2D Examples} & \multirow{2.5}{*}{NFE} & \multicolumn{2}{c}{Mixture of 5-Gaussian} & \multicolumn{2}{|c}{Two Moon} & \multicolumn{2}{|c}{Checker Board} \\
        \cmidrule(lr){3-4} \cmidrule(lr){5-6} \cmidrule(lr){7-8}
        ~ & ~ & Wasserstein$^2$ & Curvature & Wasserstein$^2$ & Curvature & Wasserstein$^2$ & Curvature \\
        \midrule
        Rectified Flow~\cite{liu2022flow} & 100 & 0.5421 & 0.0316 & 0.1006 & 0.0111 & 0.3900 & 9.1946 \\
        OT-CFM~\cite{tong2024improvinggeneralizingflowbasedgenerative} & 100 & 0.6582 & 0.0104 & 0.1074 & 0.0020 & 0.3188 & 0.1741 \\
        MeanFlow~\cite{geng2025mean} & 1 & 0.7612 & 0.9170 & 0.1233 & - & 0.9170 & - \\
        COT-FM (Ours) & 100 & \textbf{0.1995} & \textbf{0.0084} & \textbf{0.0266} & \textbf{0.0016} & \textbf{0.2550} & \textbf{0.1505} \\
        \bottomrule
    \end{tabular}
    }
    \caption{\textbf{2D synthetic data distributions.} COT-FM outperforms all baselines in both Wasserstein distance and curvature.}
    \label{tab:toy_example}
\end{table*}

\subsection{Optimizing FM Models}
The optimization stage follows the standard FM training procedure.  As detailed in Algorithm~\ref{algo:train-fm}, we compose a minibatch by randomly sampling source-target pairs based on pre-computed cluster-wise OT maps $\{\pi_k\}_{k=1}^K$.  First, we randomly draw cluster index $k$ with a probability proportional to the cluster size $\frac{|\mathcal{C}_k|}{\sum_{j=1}^K |\mathcal{C}_j|}$, followed by sampling a source-target sample pair with the corresponding OT map: $(\bm{x}_{0}, \bm{x}_{1}) \sim \pi_k$.  Secondly, we sample time step $t$ from a uniform distribution $\mathcal{U}(0, 1)$, and calculate the linear interpolation $\bm{x}_t=(1-t)\bm{x}_0 + t \bm{x}_1$.  Lastly, the FM model is optimized to regress the constructed vector field via the conditional flow-matching loss $\mathcal{L}_{\textsf{CFM}}$ (Eqn.~\ref{eqn:cfm_loss}).

\subsection{Alternating Optimization and Sampling} \label{sec:alternative_training}
Starting with a pre-trained FM model, \method{} alternates between refining cluster-wise target vector field and updating the FM model.  Empirically, we find that the model performance converges within a small number of alternation rounds.  Since \method{} does not alter the FM architecture, it preserves the standard inference procedure. The only modification lies in the initialization step: whereas the original FM model samples an initial source point from a single global source distribution, \method{} first samples a cluster index $k$ and then draws the initial sample from the corresponding source distribution $p_{0,k}$. The cluster index $k$ is sampled according to the probability $\frac{|\mathcal{C}_k|}{\sum_{j=1}^K |\mathcal{C}_j|}$, where $|\mathcal{C}_k|$ denotes the number of samples in cluster $k$, for a multinomial distribution.  
Other sampling methods are described in Section~\ref{abl:sample_method}. The full generation process is summarized in Algorithm~\ref{algo:sample}.

\section{Experiments}
\label{sec:experiments}

\begin{figure}[t]
    \centering
    \includegraphics[width=\linewidth]{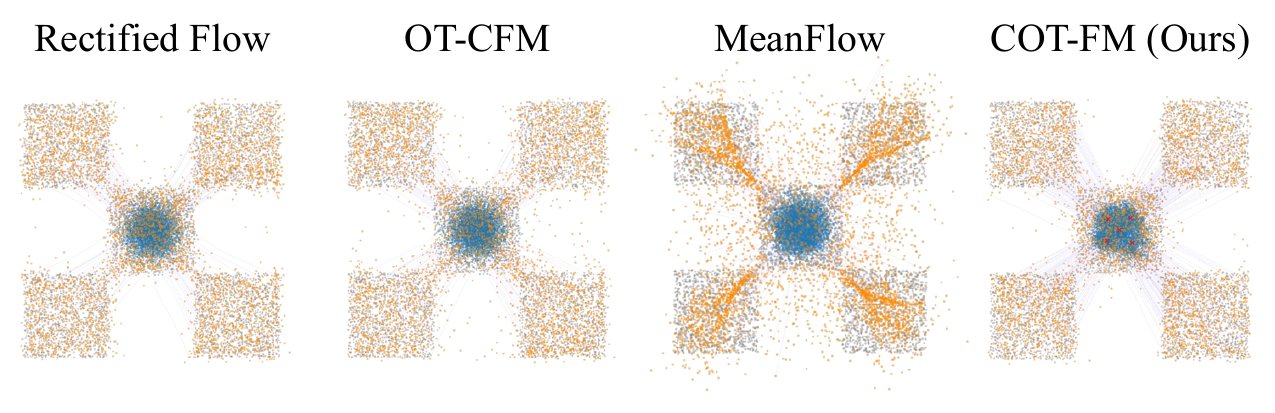}
    \caption{\textbf{Visualization of 2D checkerboard data under different methods.} Blue points denote the source distribution, gray points the target distribution, and orange points the generated samples.}
    \label{fig:checker_board}
\end{figure}

We evaluate \method{} under the following setups: unconditional 2D point cloud generation, unconditional and conditional image generation on CIFAR-10 and ImageNet, and text-conditional robotic manipulation on LIBERO. In addition, we perform an extensive ablation study to assess the contribution of each model component.

\subsection{Unconditional 2D Point Cloud Generation}
To preliminarily validate our idea, we design a simple benchmark for unconditional 2D point cloud generation.  We consider three types of data distributions: a mixture of 5 Gaussians, two moons, checkerboard.

We compare against the following baseline methods: (1) Rectified Flow~\cite{liu2022flow}, which uses random couplings; (2) OT-CFM~\cite{tong2024improvinggeneralizingflowbasedgenerative}, which uses batch-wise optimal couplings; (3) MeanFlow~\cite{geng2025mean}, which uses random couplings and learns an average velocity field to skip sampling.  All baseline models use a standard Gaussian source distribution $p_0(\bm{x}) = \mathcal{N}(\bm{x}; \bm{0}, \bm{I})$.  For OT-CFM and \method{}, we use the same backbone FM model as Rectified Flow.  We apply the K-means algorithm to segment data samples into 5 clusters, and perform two alternation cycles during training.

Following its typical setup, MeanFlow generates data with $1$ sampling step, while others use $100$ steps.  To evaluate model performance, we consider two metrics: \textit{Wasserstein distance} and \textit{curvature}.  The former measures how close the generated distribution is to the target distribution, and the latter quantifies the straightness of the learned flows.

We present our quantitative results in Table~\ref{tab:toy_example}, \method{} outperforms all baselines significantly.  It achieves the smallest Wasserstein distance--$0.1995$, $0.0266$, $0.2550$ compared to $0.5421$, $0.1006$, $0.3900$, and the least curvature--$0.0084$, $0.0016$, $0.1505$ compared to $0.0104$, $0.0020$, $0.1741$, of the second best method on the four types of data distributions.  We show qualitative results in Fig.~\ref{fig:teaser} and Fig.~\ref{fig:checker_board}.  Our generated 2D point cloud is much closer to the original data distributions than others.  Notably, we observe that the learned flows of MeanFlow remain curved and its generated samples deviate from the data distributions.  These results support our proposition that cluster-wise OT enhances generation quality of FM models, and shortcut models only accelerate but do not enhance quality of data generation.  Additional details are provided in the Supplementary Material.

\begin{table}[t]
\centering
\begin{adjustbox}{width=\linewidth}
\begin{tabular}{l|cccc}
    \toprule
    Dataset: CIFAR-10 & 1-step & 2-step & 10-step & 50-step \\
    \midrule
    Rectified Flow~\cite{liu2022flow} &  &  &  &  \\
    \quad random coupling & 378.0 & 173 & 12.6 & 4.45  \\
    \quad + clustering & 296.0 & 107 & 10.1 & 4.19 \\
    \quad OT-CFM~\cite{tong2024improvinggeneralizingflowbasedgenerative} & 226.0 & 82.2 & 10.6 & 4.78 \\
    \quad \method{} (Ours) & \textbf{205.0} & \textbf{59.1} & \textbf{8.23} & \textbf{3.97} \\
    \midrule
    MeanFlow~\cite{geng2025mean} &  &  &  &  \\
    \quad random coupling & 2.92 & 2.88 &  - & - \\
    \quad \method{} (Ours) & \textbf{2.60} & \textbf{2.53} & - & - \\
    \bottomrule
\end{tabular}
\end{adjustbox}
\caption{\textbf{Unconditional image generation on CIFAR-10~\cite{krizhevsky2009learning}.} \method{} improves generation quality across different flow matching baselines and sampling steps. Measured in FID (lower is better).}
\label{tab:cifar10}
\end{table}

\subsection{Unconditional Image Generation}

Next, we evaluate \method{} on the more challenging task of unconditional image generation on CIFAR-10~\cite{krizhevsky2009learning}, which contains 50,000 training images of resolution $32 \times 32$.  We compare \method{} against three baseline methods: Rectified Flow, OT-CFM, and MeanFlow. Given that \method{} is model-agnostic, we integrate it with Rectified Flow and MeanFlow, and evaluate each resulting model independently against its original counterpart.

\begin{figure}
    \centering
    \includegraphics[width=\linewidth]{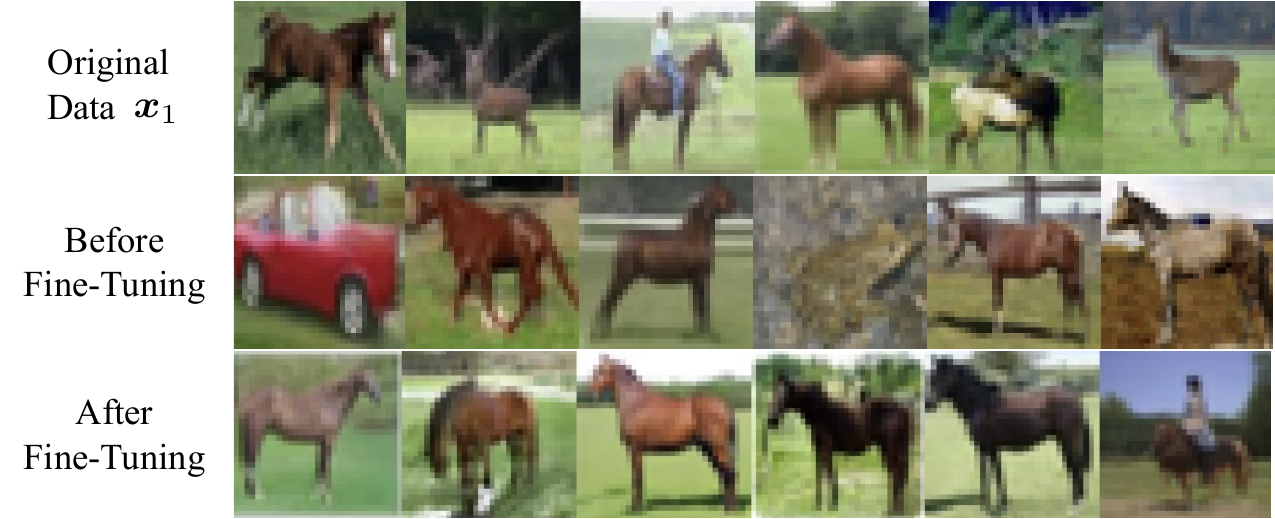}
    \caption{Visualization of a CIFAR-10 cluster. The first row shows representative images from the cluster. Using these images, we apply reverse ODE to estimate the $\bm{\mu}_{0,k}$ and $\bm{\Sigma}_{0,k}$ (Fig.~\ref{fig:method}, Stage 1). The second row shows images generated by sampling noise from this estimated distribution. The last row presents the results after finetuning the pretrained flow model.}
    \label{fig:cifar10_viz_finetune}
    \vspace{-5pt}
\end{figure}

For clustering images, we first encode an image into a latent feature vector using a self-supervised representation learning framework--DINO~\cite{caron2021emergingpropertiesselfsupervisedvision}, then apply K-Means algorithm to segment all training images into $100$ clusters.  It has been shown that clusters derived from such self-supervised features closely correspond to image categories~\cite{wu2018unsupervised,oquab2023dinov2}. The clustering result is shown in Fig.~\ref{fig:cifar10_viz_finetune}(a).

We adopt Fréchet Inception Distance (FID)~\cite{heusel2018ganstrainedtimescaleupdate} as the evaluation metric, which measures the similarity between generated data and real data distribution in the latent feature space~\cite{szegedy2015going}.  Lower FID indicates better generation quality. See the Supplementary Material for architecture, training, and evaluation details.

\begin{figure}
    \centering
    \includegraphics[width=\linewidth]{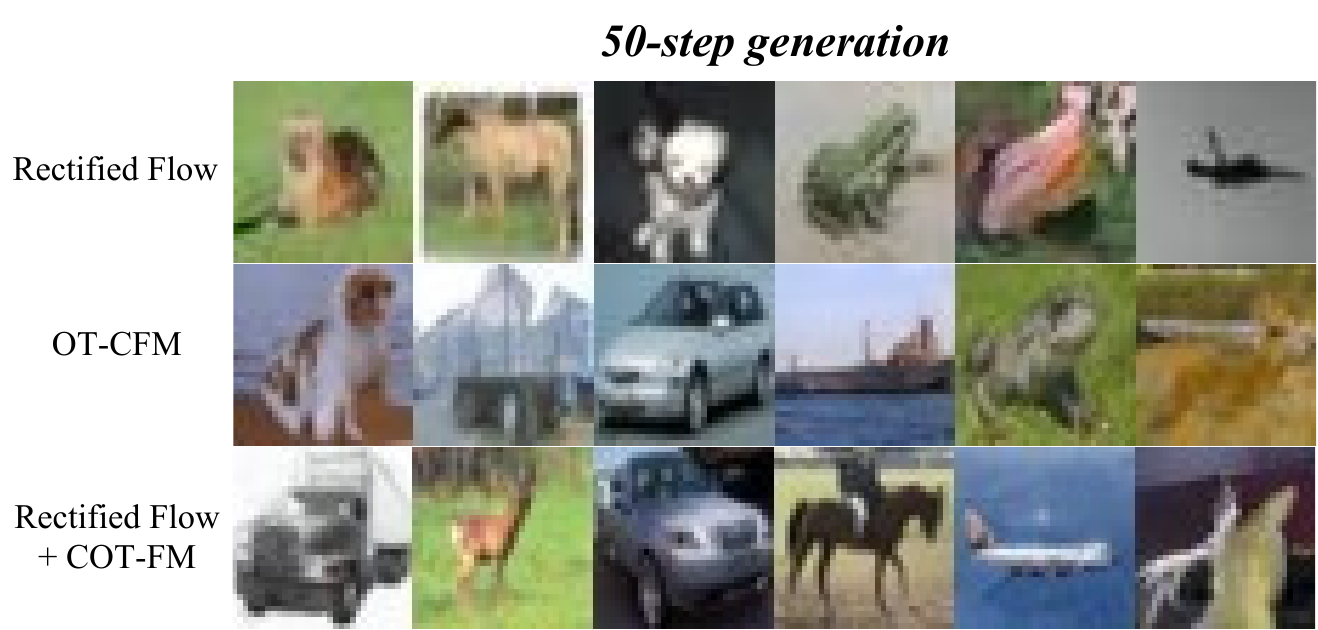}
    \caption{CIFAR-10 visualization for different flow-based methods under 50-step generation settings.}
    \label{fig:cifar10_viz}
\end{figure}

Table~\ref{tab:cifar10} summarizes the results. When using the same backbone FM model as Rectified Flow, introducing cluster-wise random coupling already yields an improvement over the original Rectified Flow without clustering, achieving an absolute performance gain of $82$ in FID.  Building on this, computing batch-wise optimal coupling within each cluster further enhances performance, yielding an additional FID reduction of 91 with a single sampling step.  Remarkably, our \method{} consistently enhances the FM model when using the same backbone as MeanFlow.  It reduces FID from $2.92$ to $2.60$ and $2.88$ to $2.53$ with one and two sampling steps. As shown in Fig.~\ref{fig:cifar10_viz}, our method produces noticeably clearer 1-step generations than Rectified Flow and improves the visual quality over OT-CFM on several object categories. For the 50-step generation setting, COT-FM still achieves a lower FID, reducing it from $4.45$ to $3.97$. These results highlight the benefit of incorporating sample clustering and optimal coupling into the training pipeline of FM models.

\begin{table}[t]
    \centering
    \resizebox{\linewidth}{!}{%
    \begin{tabular}{l|l|ccccc}
    \toprule
    Model & Method & 100 & 50 & 10 & 2 & 1 \\
    \midrule
    \multirow{2}{*}{SiT-B/2} & Rectified Flow~\cite{liu2022flow} & 5.82 & 5.86 & 8.25 & 119.57 & 264.36 \\
     & COT-FM (Ours) & \textbf{5.11} & \textbf{5.28} & \textbf{7.52} & \textbf{101.66} & \textbf{231.99} \\
    \midrule
    \multirow{2}{*}{SiT-B/4} & Rectified Flow~\cite{liu2022flow} & 8.30 & 8.39 & 11.16 & 134.99 & 276.13 \\
     & COT-FM (Ours) & \textbf{7.65} & \textbf{7.81} & \textbf{9.87} & \textbf{114.10} & \textbf{241.18} \\
    \bottomrule
    \end{tabular}
    }
    \caption{\textbf{Conditional Image Generation on ImageNet 256$\times$256.} FID ($\downarrow$) at different NFE steps.}
    \label{tab:imagenet}
\end{table}

\begin{figure}[t]
    \centering
    \begin{minipage}{\linewidth}
        \centering
        \begin{tabular}{cc}
        \centering
        \tb{@{}l|c|cc@{}}{2.0}{
            \toprule
            Dataset: LIBERO & NFE & Spatial & Long \\
            \midrule
            FLOWER~\cite{reuss2025flower} & 4 & 97.1\% & 93.5\% \\
            \midrule
            FLOWER~\cite{reuss2025flower} & 1 & 94.2\% & 87.3\% \\
            2-Rectified Flow~\cite{liu2022flow} & 1 & 95.7\% & 91.5\%\\
            \method{} (ours) & 1 & 96.1\% & \textbf{94.5}\% \\
            \bottomrule
        }
        \end{tabular}
        \captionof{table}{\textbf{Robotic manipulation on LIBERO benchmarks~\cite{liu2023libero}.} \method{} achieves single-step performance comparable to FLOWER's 4-step generation while outperforming other single-step baselines. Measured in success rate (higher is better).}
        \label{tab:robot}
    \end{minipage}
    
    \vspace{30pt}
    
    \begin{minipage}{\linewidth}
        \centering
        \includegraphics[width=\linewidth]{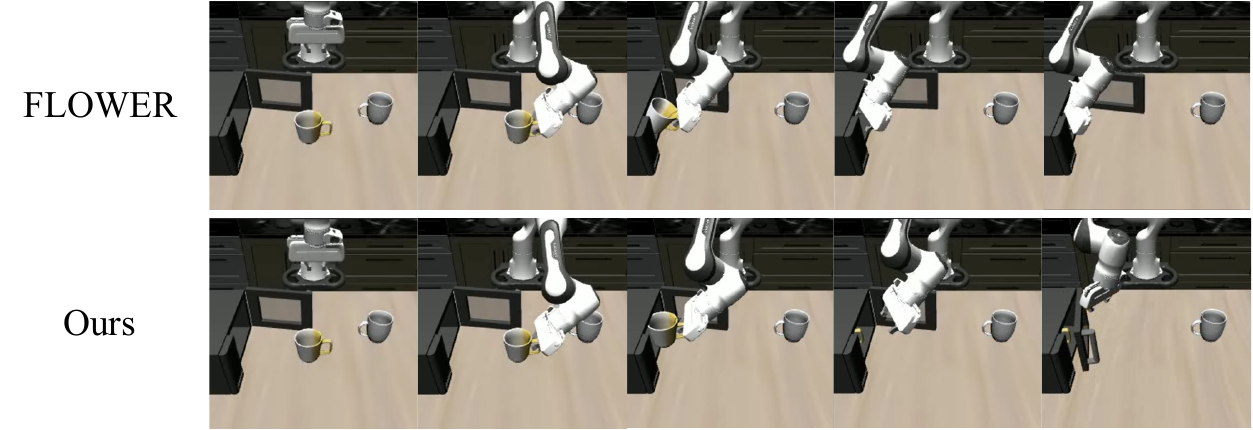}
        \caption{Comparison of two trajectories under the same initial setting in LIBERO Long. Our method successfully places the mug into the microwave and closes the door, while FLOWER fails to insert the mug smoothly and gets stuck inside the microwave.}
        \label{fig:robot}
    \end{minipage}
    \vspace{-15pt}
\end{figure}

\subsection{Conditional Image Generation}
We further evaluate \method{} on conditional image generation using ImageNet~\cite{deng2009imagenet} at 256$\times$256 resolution. We adopt SiT-B/2 and SiT-B/4~\cite{ma2024sit} as backbone architectures. In the conditional setting, we group training samples by class labels, treating each class as a cluster for constructing cluster-wise source distributions. For comparison, we train COT-FM by continuing from a pre-trained 1-RF model.

As shown in Table~\ref{tab:imagenet}, \method{} consistently improves FID across all NFE settings for both architectures. Notably, the improvement is more pronounced in the low-NFE regime, where straighter transport paths are particularly beneficial. These results demonstrate that our method scales effectively
to high-resolution, class-conditional benchmarks without modification to the core algorithm.

\subsection{Text-conditional Robotic Action Generation}
We evaluate \method{} on text-conditional robotic action generation using the LIBERO benchmark~\cite{liu2023libero}, which provides tabletop manipulation tasks paired with natural-language instructions. This simulation benchmark consists of tabletop manipulation tasks involving a Franka robot arm~\cite{gaz2019dynamic} and 3D object assets. The goal is to generate robot actions that accomplish the manipulation task conditioned on textual instructions. LIBERO comprises multiple task suites; we evaluate on LIBERO-Spatial and LIBERO-Long. The former emphasizes spatial variations, while the latter focuses on long-horizon task execution.

We compare against two baselines: (1) FLOWER~\cite{reuss2025flower}, a text-conditioned FM policy trained with random couplings, and (2) Reflow~\cite{liu2022flow}, which employs rectified couplings between source and generated samples.  Both \method{} and 2-Rectified Flow use the same backbone as FLOWER for fair comparison.  We evaluate performance using the \textit{success rate}, defined as the proportion of successful task completions across all evaluation trials.

Results are summarized in Table~\ref{tab:robot}. Using only a single sampling step, \method{} outperforms all baseline methods, achieving absolute gains of 0.4\% on LIBERO-Spatial and 3\% on LIBERO-Long over the second-best method. Remarkably, \method{} with one sampling step surpasses FLOWER with four steps. This improvement arises because cluster-wise optimal couplings produce straighter vector fields than random couplings, enabling the FM model to learn more accurate flows. See Fig.~\ref{fig:robot} for rollout visualizations.

\subsection{Ablation Analysis}

We conduct ablation studies on CIFAR-10 to analyze design choices of \method{}. All experiments use 1-RF with 50-step generation unless otherwise specified.

\paragraph{Alternating Iterations.} We examine how many alternating optimization cycles are needed. Starting from a pre-trained Rectified Flow model, we alternate between training the flow matching model with cluster-wise optimal transport and refining source distributions through reverse ODE integration. Table~\ref{tab:ablation}(a) shows that each iteration progressively reduces FID: 4.45 for the initial model improves to 4.23, 3.97, and 4.17 after 1, 2, and 3 iterations respectively. The diminishing returns suggest that 2 iterations suffice for best performance.

\paragraph{Test Set Generalization.}
Since \method{} modifies source distributions based on training data clusters, we verify it does not overfit. Table~\ref{tab:ablation}(b) compares FID on training versus held-out test images. It is expected that test dataset has higher FID than train dataset due to only 10000 images in CIFAR-10 test dataset. Both Rectified Flow baseline and \method{} show consistent performance between train and test sets—4.45, 8.55 for Rectified Flow and 3.97, 8.19 for \method{}—confirming that cluster-wise source distributions generalize without degrading quality on unseen data.

\paragraph{Different Cluster Sample Probabilities.}~\label{abl:sample_method}
Table~\ref{tab:ablation}(c) compares FID using different methods to sample cluster labels. We show that using the proportional to cluster size as a sampling probability achieves better performance than uniform sampling. We argue that this is because aligning the dataset distribution requires assigning higher sampling frequency to larger clusters, which help maintain data balance and improve FID.

\begin{table}[t]
\centering
\small

\begin{adjustbox}{max width=\linewidth}
\begin{tabular}{lc@{\hspace{2.5em}}|lc}
    \toprule
    \multicolumn{2}{c|}{\textbf{(a) Alternating iterations}} & \multicolumn{2}{c}{\textbf{(b) Test set generalization}} \\
    \midrule
    Method & FID & Method & FID \\
    \midrule
    Rectified Flow (0 iter.) & 4.45 & Rectified Flow (train) & 4.45 \\
    \method{} (1 iter.) & 4.23 & Rectified Flow (test) & 8.55 \\
    \method{} (2 iter.) & \textbf{3.97} & \method{} (train) & \textbf{3.97} \\
    \method{} (3 iter.) & 4.17 & \method{} (test) & \textbf{8.19} \\
    \bottomrule
\end{tabular}
\end{adjustbox}

\vspace{10pt}

\setlength{\tabcolsep}{30pt}
\begin{tabular}{lc}
    \toprule
    \multicolumn{2}{c}{\textbf{(c) Different cluster sampling probabilities}} \\
    \midrule
    Method & FID \\
    \midrule
    Uniform & 4.26 \\
    Proportional to cluster size & \textbf{3.97} \\
    \bottomrule
\end{tabular}

\caption{Ablation studies on CIFAR-10 at 50-step generation. (a) Additional alternating iterations progressively reduce FID. (b) \method{} maintains consistent performance on held-out test data. (c) Size-proportional cluster sampling outperforms uniform sampling.}
\label{tab:ablation}
\end{table}

\section{Conclusions}
We present \method{}, a general framework that reshapes Flow Matching probability paths for faster and more reliable generation. By clustering data and assigning each cluster its own reversed source distribution, COT-FM straightens trajectories normally distorted by random or batch-wise couplings. This yields a more accurate vector field without altering model architectures or training pipelines. As a plug-and-play method, COT-FM consistently accelerates sampling while improving quality across 2D synthetic data, image generation, and robotic manipulation.

\paragraph{Limitations and Future Work.}
While COT-FM improves flow straightness and sampling efficiency across diverse domains, challenges remain in scaling and generalizing the approach. The reverse ODE step can become inefficient as sample size grows, and our approach depends on clustering quality. The behavior of cluster-wise transport—particularly how local OT captures global structure and why alternating refinement converges within one or two cycles—remains not fully understood. Exploring more scalable clustering strategies and lighter source-distribution estimators offers promising directions for extending COT-FM.

\clearpage
{
    \small
    \section*{Acknowledgment}
    This work was supported in part by the AMD--ITRI Joint Laboratory, which provided MI300X high-performance computing resources and technical support for the execution and validation of this research. This work was also supported by the AMD University Program AI \& HPC Cluster. We further acknowledge Kuo-Guang Tsai for his technical support on the AMD system cluster infrastructure. This research was also supported by the National Science and Technology Council (NSTC), Taiwan, under Grants 114-2222-E-002-008, 114-2221-E-002-182-MY3, 113-2221-E-002-201, and 115-2634-F-002-001.
    \bibliographystyle{ieeenat_fullname}
    \bibliography{bibs/main}
}

\clearpage
\begin{center}
    \LARGE{\textbf{Supplementary Material}}
\end{center}

\renewcommand{\thesection}{\Alph{section}}
\renewcommand{\theHsection}{\Alph{section}}
\setcounter{section}{0}
\section{Extension to Conditional Generation}
\label{sec:conditional}

The COT-FM framework naturally extends to conditional generation settings. In unconditional generation, we use unsupervised clustering to partition the target distribution. In conditional generation, the conditions themselves serve as natural cluster labels. For instance, in text-to-image generation, images corresponding to the same text prompt (e.g., "cat") naturally form a cluster. Similarly, in robot manipulation tasks, different observations correspond to different action distributions, providing an implicit cluster.

The key difference is that conditional generation may encounter unseen conditions at test time, such as input observation for robot policy. Unlike unconditional generation where we can precompute mean $\bm{\mu}_k$ and covariance matrix $\bm{\Sigma}_k$ for each cluster $k$, we cannot enumerate all possible conditions in advance. To address this, we learn a conditional model $\phi$ that dynamically predicts the noise distribution for any given condition $c_k$:
\begin{equation}
\bm{x}_0 \sim \mathcal{N}(\bm{\mu}_\phi(c_k), \bm{\sigma}^2_\phi(c_k)I),
\label{eq:conditional_noise}
\end{equation}
where $\bm{\mu}_\phi(c_k)$ and $\bm{\sigma}_\phi(c_k)$ are the predicted mean and standard deviation conditioned on $c_k$. Since learning a full covariance matrix in high-dimensional space is prohibitively difficult and unstable, we approximate the covariance using only the predicted standard deviation.
This allows COT-FM to generate samples from appropriate cluster-wise noise distributions at inference time, maintaining the benefits of reduced trajectory curvature while generalizing to novel conditions.
We formulate the problem of training a conditional model $\phi$ that predicts mean $\bm{\mu}_\phi(c_k)$ and standard deviation $\bm{\sigma}_\phi(c_k)$ for a given condition $c_k$ as a reinforcement learning problem within a one-step Markov Decision Process (MDP). 

The one-step MDP is defined by a tuple $(\mathcal{S}, \mathcal{A}, R)$, where $\mathcal{S}$ denotes the state space, $\mathcal{A}$ the action space, and $R$ the reward function. Since our framework terminates after one step, at timestep $t=0$, the conditional model $\phi$ observes a state $s_0 \in \mathcal{S}$, outputs an action $a_0 \in \mathcal{A}$ and earns a reward $R(s_0, a_0)$. The objective is to maximize the expected return over all states in $\mathcal{S}$.  We adopt Proximal Policy Optimization (PPO)~\cite{schulman2017proximal} algorithm for this reinforcement learning problem.

As shown in Algorithm~\ref{algo:non-fixed-init-sources}, given an offline dataset $\mathcal{D}$ with $\{(c_k, X_{1,k})\}_{k=1}^K$, our goal is to learn a conditional distribution that best matches the distribution of $X_{1,k}$ under condition $c_k$. We therefore treat the condition $c_k$ as state $s_0$ and specify the action as drawing an initial point $\bm{x}_0$ according to $c_k$ 
\begin{align}
    \bm{x}_0 \sim \mathcal{N}(\bm{\mu}_\phi(c_k), \bm{\sigma}^2_\phi(c_k)I),
\end{align}
as $a \in \mathcal{A}$.
The sampled $\bm{x}_0$ is then propagated through the ODE to obtain a terminal point $\hat{\bm{x}}_1$.
We define the reward for this transition as 
\begin{align}
R(s_0, a_0) = \operatorname{exp}(-\operatorname{MSE}(\hat{\bm{x}}_1, \bm{x}_1)).
\end{align}
where $\operatorname{MSE}$ denotes mean squared error between the generated and target terminal points.
Because the MDP contains only a single transition, the value equals the reward:
\begin{align}
V = \operatorname{exp}(-\operatorname{MSE}(\hat{\bm{x}}_1, \bm{x}_1)),
\end{align}
We compute the advantage as
\begin{align}
\hat{A} = V- \mathbb{E}_{p_k}[V],
\end{align}
where $p_k$ is the distribution from condition $c_k$.
We then use these data to update $\phi$ with PPO.

During finetuning of the pretrained flow model in Algorithm~\ref{algo:non-fixed-train-fm} and during sampling in Algorithm~\ref{algo:non-fixed-sample}, the procedure follows our original algorithm, except that the predefined $(\bm{\mu}_k, \bm{\sigma}_k)$ are replaced with the conditional outputs $(\bm{\mu}_\phi(c_k),\bm{\sigma}_\phi(c_k))$.  
\begin{algorithm}[t]
\caption{Training conditional model for non-fixed condition in robot policy}
\label{algo:non-fixed-init-sources}
\begin{algorithmic}[1]
\STATE \textbf{Input:} Trained FM model $\bm{v}_\theta$, initialized conditional model $\phi$, offline dataset $\mathcal{D}$ with $\{(c_k, X_{1,k})\}_{k=1}^K$, batch size $M$, training steps $L$.

\FOR{$l = 1$ to $L$}
    \STATE Initialize minibatch: $B \leftarrow \{\}$
    \FOR{$m = 1$ to $M$}
        \STATE Sample $(c_k, \bm{x}_1)$ from dataset $\mathcal{D}$
        \STATE $\bm{x}_0 \sim \mathcal{N}(\bm{\mu}_\phi(c_k), \bm{\sigma}^2_\phi(c_k)I)$
        \STATE Compute log-likelihood $\log p_\phi(\bm{x}_0 \mid c_k)$
        \STATE Roll out FM: $\hat{\bm{x}}_1 \leftarrow \operatorname{ODESolver}(\bm{v}_\theta, \bm{x}_0, c_k)$
        \STATE $V \leftarrow \exp(-\operatorname{MSE}(\hat{\bm{x}}_1, \bm{x}_1))$
        \STATE $\hat{A} \leftarrow V - \mathbb{E}_{p_k}[V]$
        \STATE Append to minibatch: $B \leftarrow B \cup \{(c_k, \bm{x}_0, \hat{A}, \log p_\phi(\bm{x}_0 \mid c_k)\}$
        
    \ENDFOR
    \STATE Optimize $\phi$ according to minibatch with PPO. 
    
\ENDFOR
\STATE \textbf{Output:} Trained conditional model $\phi$.
\end{algorithmic}
\end{algorithm}

\begin{algorithm}[t]
\caption{Fine-tuning FM model for non-fixed condition in robot policy}
\label{algo:non-fixed-train-fm}
\begin{algorithmic}[1]
\STATE \textbf{Input:} FM model $\bm{v}_\theta$, conditional model $\phi$, batch size $M$, training steps $L$, offline dataset $\mathcal{D} = \{(c_k, X_{1,k})\}_{k=1}^K$.
\FOR{$l = 1$ to $L$}
    \STATE Initialize minibatch: $B \leftarrow \{\}$
    \FOR{$m = 1$ to $M$}
        \STATE Sample batch data  $(c_k, \bm{x}_1) \sim \mathcal{D}$
        \STATE $\bm{x}_0 \sim \mathcal{N}(\bm{\mu}_\phi(c_k), \bm{\sigma}^2_\phi(c_k)I)$
        \STATE Append to minibatch: $B \leftarrow B \cup \{(\bm{x}_0, \bm{x}_1)\}$
    \ENDFOR
    \STATE $\mathcal{L}_{\textsf{CFM}}(\theta) = \mathbb{E}_{t, (\bm{x}_0, \bm{x}_1) \sim B} \| \bm{v}_\theta(\bm{x}_t, t) - (\bm{x}_{1} - \bm{x}_{0}) \|_2^2$
    \STATE $\theta \leftarrow \theta - \gamma \nabla_\theta \mathcal{L}_{\textsf{CFM}}(\theta)$
\ENDFOR
\STATE \textbf{Output:} Updated FM model $\bm{v}_\theta$.
\end{algorithmic}
\end{algorithm}

\begin{algorithm}[t]
\caption{Sampling for non-fixed condition in robot policy}
\label{algo:non-fixed-sample}
\begin{algorithmic}[1] 
\STATE \textbf{Input:} Conditional model $\phi$, FM model $\bm{v}_\theta$, condition $c_k$, number of sampling steps $T$.
\STATE Draw source sample $\bm{x} \sim \mathcal{N}(\bm{\mu}_\phi(c_k), \bm{\sigma}^2_\phi(c_k) I)$
\FOR{$t=0$ to $T-1$}
    \STATE $\bm{x} \leftarrow \bm{x} + \bm{v}_\theta(\bm{x}, \frac{t}{T}) \cdot \frac{1}{T}$
\ENDFOR
\STATE \textbf{Output:} Generated target sample $\bm{x}$.
\end{algorithmic}
\end{algorithm}

\section{Computational Cost Analysis}
\label{sec:computational_cost}

\paragraph{Theoretical Analysis.}
Let $n$ be the dataset size, $b$ the batch size, and $K$ the number of clusters. Exact OT costs $O(n^3)$, while batch OT (used by OT-CFM) costs $O(b^3 \cdot (n/b)) = O(nb^2)$ per epoch~\cite{peyre2019computational}. COT-FM computes exact OT within $K$ clusters each epoch: $O((n/K)^3 \cdot K) = O(n(n/K)^2)$. Our focus is the speed--quality tradeoff: as shown in \cref{tab:cost_analysis}, COT-FM achieves better FID at matched per-epoch OT wall-clock time.

\begin{table}[htbp]
\centering
\caption{Speed--quality tradeoff comparison on CIFAR-10. OT-CFM with batch size $b=512$ and COT-FM with cluster size $n/K=500$ yield comparable per-epoch OT computation time, but COT-FM achieves significantly better FID.}
\label{tab:cost_analysis}
\resizebox{0.95\linewidth}{!}{%
\begin{tabular}{ccc|ccc}
\toprule
\multicolumn{3}{c|}{OT-CFM (Batch OT)} & \multicolumn{3}{c}{COT-FM (Ours)} \\
\midrule
$b$ & Time (sec.) & FID & $n/K$ & Time (sec.) & FID \\
\midrule
512 & 6.2 & 4.78 & 500 (50000/100) & 6.3 & \textbf{3.97} \\
\bottomrule
\end{tabular}
}
\end{table}

\paragraph{Empirical Time Breakdown.}
We measure the wall-clock time of each preprocessing stage on CIFAR-10. As shown in \cref{tab:time_breakdown}, the one-time preprocessing takes 445 seconds in total, dominated by the reverse ODE computation. The per-epoch OT computation is approximately 15 seconds.

\begin{table}[htbp]
\centering
\caption{Wall-clock time breakdown for one-time preprocessing on CIFAR-10.}
\label{tab:time_breakdown}
\begin{tabular}{lc}
\toprule
One-time Preprocessing & Time Cost (sec.) \\
\midrule
Reverse ODE & 420 \\
DINO feature extraction & 15 \\
K-means clustering & 10 \\
\bottomrule
\end{tabular}
\end{table}

\paragraph{Overhead of Reverse ODE vs.\ Training Longer.}
On CIFAR-10, the reverse ODE preprocessing takes 420 seconds, equivalent to approximately 5 training epochs. Training RF for these additional epochs increases FID from 4.45 to 4.49 due to overfitting, while OT-CFM shows no improvement---FID remains at 4.78. In contrast, COT-FM achieves FID \textbf{3.97}, confirming that the preprocessing overhead is well justified by the quality gains.

\section{Comparison with 2-Rectified Flow}
\label{sec:2rf_comparison}

2-Rectified Flow (2-RF)~\cite{liu2022flow} is a standard rectification method that also involves a two-stage pipeline: (1) train 1-RF, then (2) generate synthetic couplings via forward ODE and retrain the model. We provide a direct comparison on CIFAR-10 to demonstrate that COT-FM is complementary to 2-RF.

As shown in \cref{tab:2rf}, applying COT-FM on top of 2-RF yields additional gains: 2-RF achieves 12.21 FID at 1-step generation, while 2-RF + COT-FM achieves \textbf{10.91} FID. This demonstrates that the two approaches address different aspects of the problem---2-RF straightens trajectories via retraining on synthetic couplings, while COT-FM further reduces path crossings through cluster-wise optimal transport---and can be combined for cumulative improvement.

\begin{table}[htbp]
\centering
\caption{1-step FID ($\downarrow$) comparison on CIFAR-10 between 2-Rectified Flow and 2-Rectified Flow enhanced with COT-FM.}
\label{tab:2rf}
\begin{tabular}{lc}
\toprule
Method & FID ($\downarrow$) \\
\midrule
2-Rectified Flow~\cite{liu2022flow} & 12.21 \\
2-Rectified Flow~\cite{liu2022flow} + COT-FM (Ours) & \textbf{10.91} \\
\bottomrule
\end{tabular}
\end{table}

\section{Additional Metrics and Ablation on $K$}
\label{sec:additional_metrics}

We report FID, Inception Score (IS), Precision, and Recall on CIFAR-10 with 50 NFE steps in \cref{tab:additional_metrics}. COT-FM improves FID, IS, and Recall compared to Rectified Flow while maintaining the same Precision, indicating that COT-FM achieves broader coverage of the target distribution without sacrificing sample fidelity.

\begin{table}[htbp]
\centering
\caption{Additional metrics on CIFAR-10 (50-step) with $K=100$.}
\label{tab:additional_metrics}
\resizebox{\linewidth}{!}{
\begin{tabular}{lcccc}
\toprule
Method & FID $\downarrow$ & IS $\uparrow$ & Precision $\uparrow$ & Recall $\uparrow$ \\
\midrule
Rectified Flow~\cite{liu2022flow} & 4.45 & 9.11 & 0.78 & 0.65 \\
COT-FM (Ours) & \textbf{3.97} & \textbf{9.36} & 0.78 & \textbf{0.67} \\
\bottomrule
\end{tabular}
}
\end{table}

We also ablate the number of clusters $K$ in \cref{tab:ablation_k}. The results show that COT-FM is robust to the choice of $K$: even with as few as $K=5$ clusters, COT-FM (FID 4.56) outperforms OT-CFM (FID 4.78). Performance peaks around $K=100$ and slightly degrades at $K=120$, suggesting that overly fine-grained clustering can reduce the number of samples per cluster and diminish the effectiveness of intra-cluster OT. In practice, we recommend applying the elbow method~\cite{thorndike1953belongs} to select $K$.

\begin{table}[htbp]
\centering
\caption{Ablation on number of clusters $K$ (CIFAR-10, 50-step FID $\downarrow$).}
\label{tab:ablation_k}
\begin{tabular}{lcccc}
\toprule
Method & $K=5$ & $K=50$ & $K=100$ & $K=120$ \\
\midrule
COT-FM & 4.56 & 4.10 & \textbf{3.97} & 4.06 \\
\bottomrule
\end{tabular}
\end{table}

\section{Discussion on Clustering Quality}
\label{sec:clustering_quality}

A natural concern is whether COT-FM is sensitive to the quality of the clustering. We argue that the method degrades gracefully and is robust in practice for the following reasons.

First, $K=1$ reduces to the full-dataset setting where exact OT is intractable, necessitating batch OT as in OT-CFM. Any $K > 1$ improves over this baseline because cluster-wise OT still reduces path crossings compared to batch OT, even with imperfect cluster assignments. As demonstrated in the ablation study (\cref{sec:additional_metrics}), even $K=5$ outperforms OT-CFM.

Second, for conditional generation tasks such as robot manipulation, COT-FM does not require explicit clustering. Instead, the conditional model learns source distributions directly from task embeddings, as described in \cref{sec:conditional}.

Third, for unconditional generation, self-supervised features from DINO~\cite{caron2021emergingpropertiesselfsupervisedvision,oquab2023dinov2} achieve 78.3\% $k$-NN accuracy on ImageNet, demonstrating that semantically meaningful clusters naturally emerge and are well-separated in real-world data. This is precisely the property that COT-FM relies on: as long as the clustering captures broad semantic structure, the intra-cluster OT assignment will produce straighter transport paths than global batch OT.

\section{Convergence of Alternating Optimization}
\label{sec:convergence}

\begin{figure}[htbp]
    \centering
    \includegraphics[width=\linewidth]{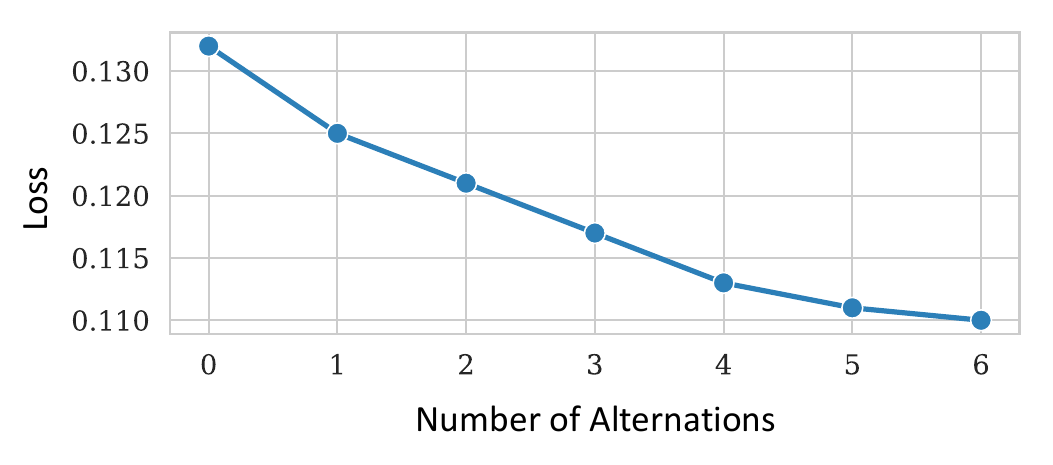}
    \caption{Training loss across alternating optimization iterations, showing monotonic decrease.}
    \label{fig:loss_convergence_supp}
\end{figure}

Table~4(a) in the main paper shows that FID improves across iterations of the alternating optimization: 4.45 (0 iterations) $\rightarrow$ 4.23 (1 iteration) $\rightarrow$ 3.97 (2 iterations), with diminishing returns at iteration 3. \cref{fig:loss_convergence_supp} shows that the training loss decreases monotonically across iterations.

Intuitively, the convergence behavior is stable because the reverse ODE produces non-intersecting paths by construction, and the subsequent cluster-wise OT reduces local transport cost within each cluster. Each iteration of the alternating procedure therefore starts from a better-aligned coupling than the previous one, leading to monotonic improvement. A formal convergence analysis remains an interesting direction for future work.

\section{Related work}


\paragraph{Diffusion Models and Flow Matching.}
Diffusion models~\cite{kingma2013auto, rezende2015variational, sohl2015deep, song2019generative, ho2020denoising, song2020score}
have emerged as a powerful framework for generative modeling. These models learn a transformation that maps a prior distribution to a target distribution by progressively adding noise to the target distribution and then learning the reverse process. The reverse process is typically formulated as a stochastic differential equation (SDE) or an ordinary differential equation (ODE)~\cite{song2020score, karras2022elucidatingdesignspacediffusionbased}.

In contrast, Flow Matching (FM)~\cite{lipman2022flow, albergo2022building}, learns a time-dependent velocity field that directly transports samples from the prior distribution to the target distribution. By modeling this velocity field, FM reduces the need to solve differential equations during training, providing a more direct and efficient way to learn the generative mapping.

\paragraph{Acceleration Strategies.}
Diffusion models and flow models sample slowly because of the recursive model evaluations required when solving the underlying ODE. To accelerate sampling while preserving generation quality, existing works can be broadly categorized into three directions: distilling knowledge from a pretrained model, learning a straighter velocity field, and directly learning a solution map of the ODE.

For distillation approaches~\cite{luhman2021knowledge, salimans2022progressive, geng2023one, sauer2024adversarial, yin2024one, zhou2024score}, the central idea is to distill the behavior of multi-step diffusion or score-based model into a few-step model. These methods supervise the student model using teacher trajectories, enabling efficient sampling without significantly sacrificing generation quality.   

For learning straighter flows, a key challenge in training FM with a straight flow path is constructing a better coupling between the source and target distributions. The original FM formulation~\cite{lipman2022flow} uses random couplings for simplicity. However, random couplings induce high curvature of flow trajectories and degrade generation quality. 
Recent works~\cite{pooladian2023multisample, tong2024improvinggeneralizingflowbasedgenerative,kornilov2024optimal,davtyan2025faster,sochopoulos2025fastflowbasedvisuomotorpolicies} address this issue by improving the couplings using Optimal Transport (OT). Another line of work explores coupling with generated samples~\cite{liu2022flow,zhu2025analyzingmitigatingmodelcollapse}. They propose an iterative training framework that couples the source distribution with the corresponding target distribution predicted by the pre-trained model.  

To directly learn the solution map of ODE~\cite{song2023consistency, kimconsistency, frans2024one, boffi2024flow, geng2025mean}, Consistency Model (CM) ~\cite{song2023consistency} learns a mapping from any point on a trajectory to its corresponding end point. Consistency Trajectory Model (CTM)~\cite{kimconsistency} generalizes this idea by learning a mapping between any two points on the trajectory. Shortcut Model~\cite{frans2024one} learns discrete shortcut transitions along the trajectory, while MeanFlow~\cite{geng2025mean} extends to a continuous formulation by learning the average velocity field between two points on the trajectory.  

\paragraph{Optimized Noise Prior.}
Many works have adopted the strategy of optimizing the noise prior to improve the performance of diffusion or flow model. In the domain of image generation, prior works~\cite{Mao_2023,mao2024lotterytickethypothesisdenoising, samuel2023generatingimagesrareconcepts} demonstrate that a specific noise prior can induce specific pattern through diffusion model. Building on this idea, subsequent works~\cite{eyring2024renoenhancingonesteptexttoimage,guo2024initnoboostingtexttoimagediffusion} enhance diffusion model performance by optimizing the initial noise using signals from reward models. Moreover, ~\cite{chen2024findfinetuninginitialnoise, li2025noisearautoregressinginitialnoise} employ reinforcement learning (RL) and Direct Preference Optimization (DPO)~\cite{rafailov2024directpreferenceoptimizationlanguage} frameworks to modify the initial noise.

In the context of robot policy learning, several works~\cite{wagenmaker2025steeringdiffusionpolicylatent,aizu2025robotmotionplanningusing} show that a better prior can lead to improved performance or reduce the number of function evaluations (NFEs) required to achieve the same performance.
\section{Experimental Configuration}
\subsection{Compute Environment}
\label{sec:compute_environment}

For the LIBERO robotics experiments, we use a single NVIDIA GeForce RTX 4090 GPU. For the CIFAR-10 and ImageNet experiments, we use the AMD AI \& HPC Clusters Taiwan, equipped with AMD Instinct MI300X GPUs (192\,GB HBM3) and AMD EPYC 9684X 96-Core Processors. GPUs within each node are interconnected via AMD Infinity Fabric (XGMI). CIFAR-10 experiments are conducted on a single MI300X GPU, while ImageNet experiments use 8 MI300X GPUs on a single node. All AMD experiments use ROCm 7.2.0.

\subsection{2D Point Cloud}
We evaluate several flow-based generative models: Rectified Flow~\cite{liu2022flow}, OT-CFM~\cite{tong2024improvinggeneralizingflowbasedgenerative}, MeanFlow~\cite{geng2025mean}, and our COT-FM on three 2D benchmarks: Mixture of 5 Gaussians, Two Moons, and a Checkerboard Grid. The source distribution is a single Gaussian with mean $(0, 0)$ and standard deviation $0.6$, while the target distribution follows the corresponding toy structure.

All methods use the same network architecture for fair comparison. We train each model using the Adam optimizer with learning rate 1e-3, batch size 512, and run 500 epochs for each task. In each iteration, source–target mini-batches are sampled, and the corresponding coupling strategy is applied. For evaluation, we measure: (1) Wasserstein distance between generated and target samples using exact 2-Wasserstein distance; (2) trajectory curvature, computed by discretizing the integrated trajectories, normalizing tangents, and measuring second-order directional change. Lower curvature corresponds to straighter probability paths.

\subsection{CIFAR-10}
We experiment with unconditional generation on CIFAR-10~\cite{krizhevsky2009learning}, where the input to the model is $32 \times 32 \times 3$ in pixel space. We evaluate  Fréchet Inception Distance (FID)~\cite{heusel2018ganstrainedtimescaleupdate} using 50K generated images. Optimal Transport is computed using the publicly released implementation from~\cite{tong2024improvinggeneralizingflowbasedgenerative}.

To ensure a fair comparison with Rectified Flow~\cite{liu2022flow}, our implementation follows their setting. We train the models for 600 epochs with a batch size of 128 using Adam optimizer with $\beta_1 = 0.9$, $\beta_2 = 0.999$, $\epsilon=1\times10^{-8}$, and no weight decay. The learning rate is set to $2\times10^{-4}$ and we apply gradient clipping with a maximum norm of 1.0. The backbone is an NCSN++-style UNet, and detailed parameters are provided in Table~\ref{tab:rf_params}. We maintain exponential moving average (EMA) weights with decay rate of 0.999999.

For comparison with MeanFlow~\cite{geng2025mean}, we adopt their experimental configuration. We train the models for 240 epochs with a batch size of 512. We used Adam optimizer with $\beta_1 = 0.9$, $\beta_2 = 0.999$ and no weight decay. The learning rate is set to $0.0001$. The UNet architecture follows~\cite{song2019generative}, with model parameters summarized in Table~\ref{tab:rf_params}. We maintain EMA weights with a decay rate of 0.99999.
\subsection{ImageNet}
We experiment with conditional generation on ImageNet~\cite{deng2009imagenet} $256 \times 256$ with full dataset. The input to the model is $32 \times 32 \times 4$ in latent space obtained from the VAE. We evaluate  Fréchet Inception Distance (FID)~\cite{heusel2018ganstrainedtimescaleupdate} using 50K generated images. Optimal Transport is computed using the publicly released implementation from~\cite{tong2024improvinggeneralizingflowbasedgenerative}.

Our implementation is slightly modified from the setting of a non-official PyTorch implementation~\cite{meanflow_pytorch} to 1-RF. The images are augmented through horizontal flipping. Both 1-RF and \method{} are trained for 190 epochs in total; \method{} is fine-tuned from a pre-trained RF model after 160 epochs, requiring only 30 additional epochs of training.
We use a batch size of 256 with the Adam optimizer with $\beta_1 = 0.9$, $\beta_2=0.95$, $\epsilon = 1 \times 10^{-8}$ and no weight decay. We set the learning rate to $1 \times 10^{-4}$ and apply gradient clipping with a maximum norm of 1.0. The backbone is Scalable Interpolant Transformers(SiT)~\cite{ma2024sit}. We maintain EMA weights with a decay rate of 0.9999. We set $\omega$ to 2.5 and 2 for SiT-B/4 and SiT-B/2, respectively. All other hyperparameters remain identical to the original configuration.
\subsection{LIBERO}
The LIBERO benchmark ~\cite{liu2023libero} uses a Franka Emika Panda robot with end-effector control and camera control. We evaluate on Spatial for spatial relationships and Long for long horizon tasks. Each task starts with 50 different initial points and we record the average success rate for each task with 3 different seeds. The hyperparameters are the same as FLOWER~\cite{reuss2025flower}, shown in Table~\ref{tab:libero_params}.

We train the reinforcement learning model over 15 passes of the offline dataset. After collecting 51,200 samples in each pass, we perform Proximal Policy Optimization (PPO)~\cite{schulman2017proximal} training for the conditional model, where condition $c_k$ contains language instructions and the current visual observations. During PPO training, the discount factor $\gamma$ is set to 0.99, the clipping parameter $\epsilon$ to 0.2, and the learning rate to $1\times10^{-5}$. Each PPO training session runs for 10 epochs.

For finetuning FLOWER, we follow the setting reported in FLOWER, as summarized in Table~\ref{tab:flower_params}. To ensure training stability, we only finetune the Action Encoders, Action Heads, and Flow Transformer.

\section{Additional Generated Results}
We present additional generated images for CIFAR-10 in this section. We use the same noise for those sampled from $\mathcal{N}(0, I)$ and same noises for those sampled from our method for consistency.

\begin{figure*}
    \centering
    \includegraphics[width=0.85\linewidth]{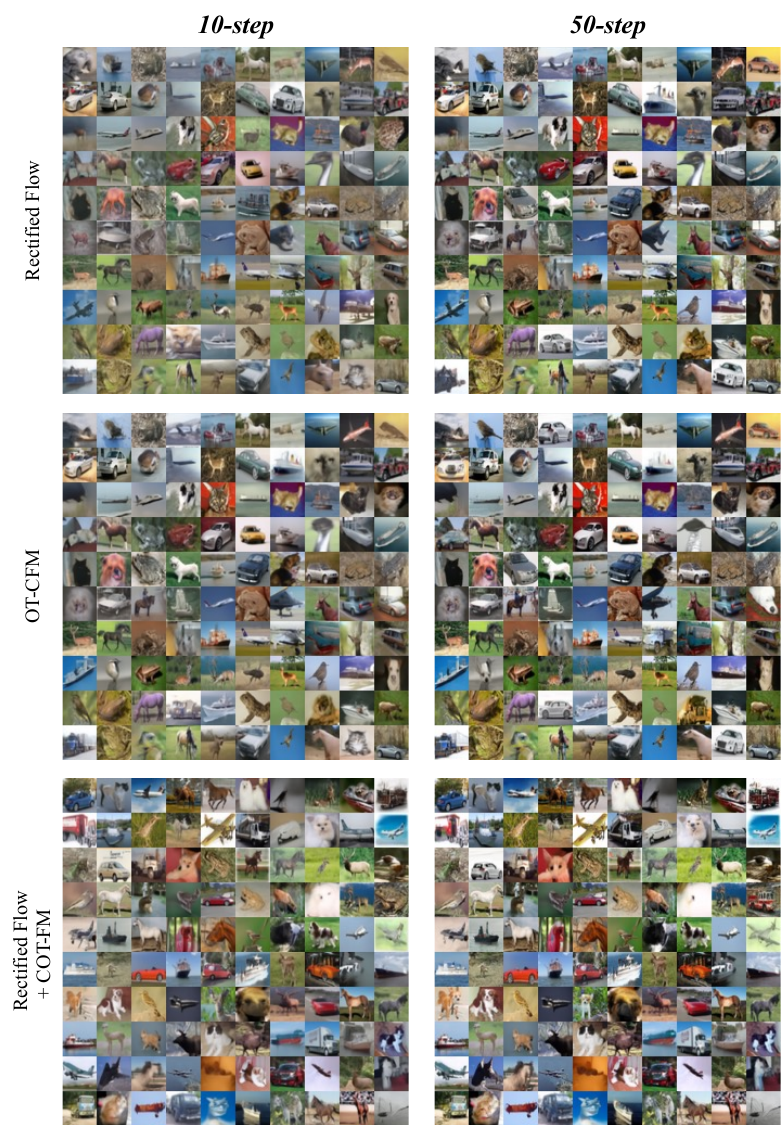}
    \caption{Visualization of CIFAR-10 for different flow-based methods under 10-step and 50-step generation settings.}
    \label{fig:cifar10_addition_viz}
\end{figure*}

\begin{figure*}
    \centering
    \includegraphics[width=0.65\linewidth]{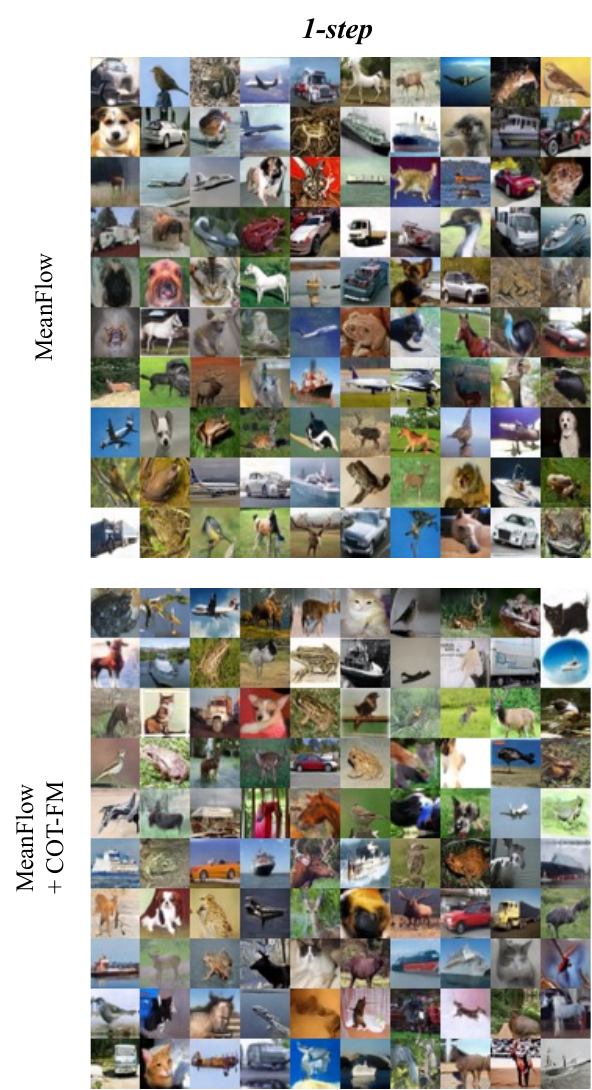}
    \caption{Visualization of CIFAR-10 for MeanFlow under 1-step generation settings.}
    \label{fig:cifar10_addition_viz_meanflow}
\end{figure*}

\begin{table}[h]
\centering
\resizebox{\linewidth}{!}{
\begin{tabular}{lcc}
\toprule
\textbf{Component} & \textbf{Rectified Flow} & \textbf{MeanFlow}\\
\midrule
Channels & 128 & 128\\
Channel multiple & (1, 2, 2, 2) & (2, 2, 2)\\
Residual blocks per resolution & 4 & 4\\
Normalization & GroupNorm & GroupNorm\\
Nonlinearity & Swish & Swish\\
Attention resolution & 16 & 16\\
Dropout & 0.15 & 0.2\\
Embedding type & Fourier & Positional \\
\bottomrule
\end{tabular}
}
\caption{UNet architecture for CIFAR-10 experiments.}
\label{tab:rf_params}
\end{table}
\begin{table}[h]
\centering
\resizebox{\linewidth}{!}{
\begin{tabular}{lc}
\hline
\textbf{Components} & \textbf{Number of Parameters} \\
\hline
ViT & 360M \\
VLM & 205M \\
Action Encoders & 3.2M \\
Action Heads & 31.8K \\
Global-AdaLN & 28.3M \\
Cond Linear Proj. & 1.0M \\
Timestep Embedder & 1.3M \\
Cond Norm & 1.0K \\
FreqEmbedder & 1.3M \\
Action Space Embedder & 1.3M \\
Flow Transformer & 339M \\
${\phi}$ (ours) & 17M \\
\hline
\textbf{Total Parameters FLOWER} & \textbf{964M} \\
\hline
\end{tabular}
}
\caption{The parameters of all model components in FLOWER.}
\label{tab:flower_params}
\end{table}

\begin{table}[h]
\centering
\resizebox{\linewidth}{!}{
\begin{tabular}{lc}
\hline
\textbf{Settings} & \textbf{LIBERO} \\
\hline
Action Space Encoders & 2-layered MLP \\
Action Space Decoders & Linear Attention \\
Number of Flow-T Layers & 18 \\
Latent Dimension & 1024 \\
Number of Heads & 16 \\
Position Embedding & 1D Rope \\
Sampling Distribution & Uniform \\
Attention Dropout & 0.1 \\
MLP Dropout & 0.1 \\
Residual Dropout & 0.1 \\
\hline
Act Seq Length & 10 \\
Denoising Steps & 4 \\
Multistep & 4 \\
Camera Views & {[Primary Static, Wrist]} \\
Use Proprio & False \\
Action Space & Delta EEF \\
Frequency & 10 \\
\hline
\end{tabular}
}
\caption{LIBERO hyperparameters of FLOWER.}
\label{tab:libero_params}
\end{table}

\end{document}